%% file: arXivtempseg.tex
\documentclass[journal]{article}
\usepackage[latin1]{inputenc}
\usepackage[T1]{fontenc}
\usepackage{graphicx}
\usepackage{bm}
\usepackage{amsmath}
\usepackage{dsfont}
\usepackage{amssymb}
\usepackage{graphicx}
\usepackage{pifont}
\usepackage{url}
\usepackage{amsfonts}
\newcommand{\tickYes}{\checkmark}
\newcommand{\tickNo}{\hspace{1pt}\ding{55}}
\newcommand{\G}{\ensuremath{\mathcal{G}}}

\input ssa_defs.tex

\title{Feature Extraction for Change-Point Detection using Stationary Subspace Analysis}
\author{Duncan A.J.~Blythe, Paul von B\"unau, Frank C.~Meinecke, \\ Klaus-Robert M\"uller}
\date{\today}

\begin{document}
\maketitle

\begin{abstract}
Detecting changes in high-dimensional time series is difficult because
it involves the comparison of probability densities that need to be estimated
from finite samples. In this paper, we present the first feature 
extraction method tailored to change point detection, which is based on 
an extended version of Stationary Subspace Analysis. We reduce the 
dimensionality of the data to the most non-stationary directions, which 
are most informative for detecting state changes in the time series. In 
extensive simulations on synthetic data we show that the accuracy of three 
change point detection algorithms is significantly increased by a prior
feature extraction step. These findings are confirmed in an application to 
industrial fault monitoring.  
\end{abstract}

\section{Introduction}

Change point detection is a task that appears in a broad 
range of applications such as biomedical signal processing \cite{Biomed2, EEGPrin, KohlmorgenKybernetics}, 
speech recognition \cite{Speech1,Speech2}, industrial process monitoring \cite{ChangePoint, Narendra}, 
fault state detection \cite{Fault1} and econometrics \cite{Econ1,Econ2}. The goal of change point 
detection is to find the time points at which a time series changes from one macroscopic state to another. 
As a result, the time series is decomposed into segments \cite{ChangePoint} of similar
behavior. Change point detection is based on finding changes in the properties of 
the data, such as in the moments (mean, variance, kurtosis) \cite{ChangePoint}, in the spectral properties \cite{Spectral}, 
temporal structure \cite{DistBased2} or changes w.r.t.~to certain patterns \cite{Pattern}.
The choice of any of these aspects 
depends on the particular application domain and on the statistical type of the changes that one
aims to detect. 

\begin{figure}[ht]
 \begin{center}
  \includegraphics[width = 87mm]{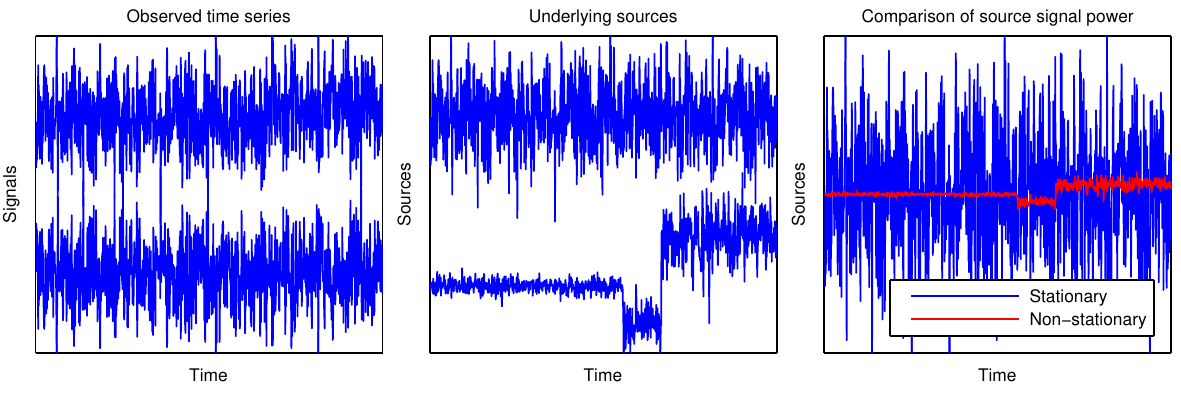}
  \caption{
  	Informative vs.~uninformative directions for change point detection. The left panel shows
	the observed bivariate time series where no pronounced changes are visible. The middle panel shows
	the two underlying sources, where one of them exhibits clearly visible changes. In the right panel, 
	we see that the stationary sources has much higher signal power than the informative non-stationary 
	sources and thus masks the presence of change points in the observed data. 
    \label{fig:tempseg_intro}
    }
 \end{center}
\end{figure}

For a large family of general segmentation algorithms, state changes are detected based on 
comparing the empirical distributions between windows of the time series 
\cite{DistBased1, DistBased4,DistBased2}. Estimating and comparing probability densities 
is a difficult statistical problem, particularly in high dimensions.
However, not all directions in the high dimensional signal space are informative for change point 
detection: often there exists a subspace in which the distribution of the data remains constant 
over time (stationary). This subspace is irrelevant for change point detection, but increases the 
overall dimensionality. Moreover, stationary components with a high signal power can make change
points invisible to the observer and also to detection algorithms. For example, there are no change 
points visible in the 
time series depicted in the left panel of Figure~\ref{fig:tempseg_intro}, even though there exists 
one direction in the two-dimensional signal space which clearly shows two change points, as it can 
be seen in the middle panel. However, the non-stationary contribution is not visible in the observed signal
because of its relatively low power (right panel). In this example, we also observe that it does
not suffice to select channels individually, as neither of them appears informative. 
In fact, in many application domains such as biomedical engineering~\cite{ZieheBio, EEGPrin, ECGBook} or geophysical 
data analysis~\cite{Geophysics}, it is most plausible that the data is generated as a mixture of 
underlying sources that we cannot measure directly. 

In this paper we show how to extract useful features for change point detection
by finding the most non-stationary directions using Stationary Subspace Analysis \cite{PRL:SSA:2009}. Even 
though there exists a wide range of feature extraction methods for classification and 
regression \cite{Guyon:2003q}, to date no specialized procedure for feature extraction or for general signal processing \cite{HaykinSig} has been proposed for change point detection. In controlled
simulations on synthetic data, we show that for three representative change point detection algorithms 
the accuracy is significantly increased by a prior feature extraction step, in particular 
if the data is high dimensional. This effect is consistent over various numbers of dimensions and strengths 
of change points. In an application to fault monitoring, where the ground truth is available, 
we show that the proposed feature extraction improves the performance and leads to a 
dimensionality reduction where the desired state changes are clearly visible. Moreover, 
we also show that we can determine the correct dimensionality of the informative subspace. 

The remainder of this paper is organized is follows. In the next Section~\ref{sec:FeEx}, we introduce 
our feature extraction method that is based on an extension of Stationary Subspace Analysis. 
Section~\ref{sec:Sim} contains the results of our simulations and in Section~\ref{sec:Real}
we present the application to fault monitoring. Our conclusions are outlined in the last Section~\ref{sec:conclusion}. 

\section{Feature Extraction for Change-Point Detection}
\label{sec:FeEx}
Feature extraction from raw high-dimensional data has been shown to be useful not only 
for improving the performance of subsequent learning algorithms on the derived 
features \cite{Guyon:2003q}, but also for understanding high-dimensional complex physical systems 
where  the relevant information is difficult to identify. In many application areas such as 
Computer Vision~\cite{Foerstner:1994q}, Bioinformatics~\cite{Saeys:2007q, Morris:2005q} and
text classification~\cite{Lewis:1991q}, defining useful features is in fact the main 
step towards successful machine learning. General feature extraction methods for classification 
and regression tasks are based on maximizing the mutual information between features 
and target \cite{Torkkolla:2003q}, explaining a given percentage of the variance in 
the dataset \cite{Schoelkopf:1998q}, choosing features which maximize the margin between 
classes \cite{Li:2006q} or selecting informative subsets of variables through enumerative search (wrapper methods) \cite{Guyon:2003q}.
However, for change-point detection no dedicated feature extraction has been proposed 
\cite{ChangePoint}. Unlike in classical supervised feature selection, where a target 
variable allows us to measure the informativeness of a feature, for change-point detection we cannot tell 
whether a feature elicits the changes that we aim to detect since there is usually 
no ground truth available. Even so, feature extraction is feasible following the 
principle that a useful feature should exhibit significant distributional changes 
over time. 
Reducing the dimensionality in a pre-processing step should be particularly beneficial 
to the change-point detection task: most algorithms either explicitly or implicitly make approximations to  
probability densities \cite{DistBased2, DistBased4} or directly compute a divergence 
measure based on summary statistics, such as the mean and covariance \cite{ChangePoint} 
between segments of the time series --- both are hard problems whose sample complexities 
grow exponentially with the number of dimensions. 

As we have seen in the example presented in Figure~\ref{fig:tempseg_intro}, selecting
channels individually (univariate approach) is not helpful or may lead to suboptimal
features. The overall data may be non-stationary notwithstanding the fact that each dimension 
seems stationary. Moreover, a single non-stationary source may be expressed across a large
number of channels. It is therefore more sensible to estimate a linear projection of the 
data which contains as much information relating to change points as possible. In this paper, 
we demonstrate that finding the projection to the most non-stationary direction
using Stationary Subspace Analysis significantly increases the performance of 
change-point detection algorithms.

In the remainder of this section, we first review the SSA algorithm and show 
how to extend it towards finding the most non-stationary directions.
Then we show that this approach corresponds to finding the projection that is  
most likely to be non-stationary in terms of a statistical hypothesis test. 

\subsection{Stationary Subspace Analysis}

Stationary Subspace Analysis \cite{PRL:SSA:2009} factorizes a multivariate 
time series $x(t) \in \R^D$ into stationary and non-stationary sources according to the linear 
mixing model,  
\begin{equation}
  x(t) = A {\mathbf s}(t) = \begin{bmatrix} A^{\s} & A^{\n} \end{bmatrix}
  \begin{bmatrix} s^{\s}(t) \\  s^{\n}(t) \end{bmatrix},
\label{eq:mixing_model}
\end{equation}
where $s^\s(t)$ are the $d_s$ stationary sources,  
$s^\n(t)$ are the $d_n$ ($d_n+d_s = D$) non-stationary sources and $A$ is an unknown
time-constant invertible mixing matrix. The spaces spanned by the columns of the mixing 
matrix $A^{\s}$ and $A^{\n}$ are called  the $\s$- and $\n$-spaces respectively. Note that in contrast 
to Independent Component Analysis (ICA)~\cite{ICABook}, there is no independence 
assumption on the sources $s(t)$. 

The aim of SSA is to invert the mixing model (Equation~\ref{eq:mixing_model}) given only 
samples from the mixed sources $x(t)$, i.e.~we want to estimate the demixing matrix 
$\hat{B}$ which separates the stationary from the non-stationary sources. 
Applying $\hat{B}$ to the time series $x(t)$ yields the estimated stationary and 
non-stationary sources $\hat{s}^\s(t)$ and $\hat{s}^\n(t)$ respectively,
\begin{align}
\label{eq:applying_solution}
	\begin{bmatrix} \hat{s}^\s(t) \\ \hat{s}^\n(t) \end{bmatrix}
	=
	\hat{B} x(t)
	=
	\begin{bmatrix} \hat{B}^\s \\ \hat{B}^\n \end{bmatrix} x(t)
	=
	\begin{bmatrix} \hat{B}^\s A^\s & \hat{B}^\s A^\n \\ \hat{B}^\n A^\s &  \hat{B}^\n A^\n \end{bmatrix}
	\begin{bmatrix} \hat{s}^\s(t) \\ \hat{s}^\n(t) \end{bmatrix} .
\end{align}
The submatrices $\hat{B}^\s \in \R^{d_s\times D}$ and $\hat{B}^\n \in \R^{(d_n)\times D}$ of the estimated
demixing matrix $\hat{B}$ project to the estimated stationary and non-stationary sources and are called
\s-projection and \n-projection respectively. The estimated mixing matrix $\hat{A}$ is the inverse of
the estimated demixing matrix, $\hat{A} = \hat{B}^{-1}$.

The inverse of the SSA model (Equation~\ref{eq:mixing_model}) is not unique: given one demixing
matrix $\hat{B}$, any linear transformation \textit{within} the two groups of estimated sources leads
to another valid separation, because it leaves the stationary resp.~non-stationary nature of the sources
unchanged. 
But also the separation into \s- and \n-sources itself is not unique: adding stationary components to
a non-stationary source leaves it non-stationary, whereas the converse is not true. That is, the
\n-projection can only be identified up to arbitrary contributions from the stationary sources.
Hence we cannot recover the true \n-sources, but only the true \s-sources (up to linear transformations).
Conversely, we can identify the true \n-space (because the \s-projection is orthogonal to it) but 
not the true \s-space. However, in order to extract features for change-point detection, our aim 
is not to recover the true non-stationary sources, but instead the \textit{most} non-stationary ones. 

An SSA algorithm depends on a definition of stationarity, that the \s-projection aims to satisfy. 
In the SSA algorithms~\cite{PRL:SSA:2009, HarKawWasBun10SSA},  a time series $X_t$ is considered
  stationary if its mean and covariance is constant over time, i.e~
\begin{align*}
  \E[X_{t_1}] & = \E[X_{t_2}] \\
  \E[X_{t_1} X_{t_1}^\top] & = \E[X_{t_2} X_{t_2}^\top],
\end{align*}
for all pairs of time points $t_1, t_2 \in \NN_0$. This is a variant of weak stationarity~\cite{Pri83Spectral} 
where we do not take time structure into account. 
Following this concept of stationarity, the SSA algorithm~\cite{PRL:SSA:2009} 
finds the \s-projection $\hat{B}^\s$ that minimizes the difference between the first two moments
of the estimated \s-sources $\hat{s}^\s(t)$ across epochs of the time series, since we cannot
estimate the mean and covariance at a single time point. Thus we divide the samples from $x(t)$
into $n$ non-overlapping epochs defined by the index sets $\mathcal{T}_1, \ldots, \mathcal{T}_n \subset \NN_0$
and estimate the epoch mean and covariance matrices,
\begin{align*}
\emu_i = \frac{1}{|T|} \sum_{t \in \mathcal{T}_i} x(t) \hspace{0.5cm} \text{ and } \\  \hspace{0.5cm}
\esi_i = \frac{1}{|T|-1} \sum_{t \in \mathcal{T}_i} \left( x(t)-\emu_i \right)\left( x(t)-\emu_i \right)^\top ,
\end{align*}
respectively for all epochs $1 \leq i \leq n$. Given an \s-projection, the epoch mean and 
covariance matrix of the estimated \s-sources in the $i$-th epoch are
\begin{align*}
	\emu^\s_i = \hat{B}^\s \emu_i \hspace{0.5cm} \text{ and } \hspace{0.5cm}
	\esi^\s_i = \hat{B}^\s \esi_i  ( \hat{B}^\s )^\top .
\end{align*}
The difference in the mean and covariance matrix between two epochs is measured using 
the Kullback-Leibler divergence between Gaussians. The objective function is the sum of the difference 
between each epoch and the average epoch. Since the \s-sources can only be determined up to an 
arbitrary linear transformation and since a global translation of the data does not change the difference
between epoch distributions, 
without loss of generality we center and whiten\footnote{A whitening transformation is a basis 
transformation $W$ that sets the sample covariance matrix to the identity. 
It can be obtained from the sample covariance matrix $\hat{\Sigma}$ as $W = \hat{\Sigma}^{-\frac{1}{2}}$.
}
the data such that average epoch's mean and covariance matrix are,
\begin{align}
\label{eq:avg_epoch}
	\frac{1}{N} \sum_{i=1}^N \emu_i = 0 \hspace{0.5cm} \text{ and } \hspace{0.5cm} \frac{1}{N} \sum_{i=1}^N \esi_i = I .
\end{align}
Moreover, we can restrict the search for the true \s-projection to the set of
matrices with orthonormal rows, i.e.~$\hat{B}^\s (\hat{B}^\s)^\top = I$. Thus 
the optimization problem becomes, 
\begin{align}
	\hat{B}^s & =
	\argmin_{B B^\top = I} \; \sum_{i=1}^N \KLD \Big[ \Gauss(\emu^\s_i,\esi^\s_i) \; \Big|\Big| \; \Gauss(0,I) \Big] \notag \\
	& = \argmin_{B B^\top = I} \; \sum_{i=1}^N \left(
			- \log\det\esi^\s_i
			+ (\hat{\mu}^\s_i)^\top \emu^\s_i 	\right) ,
\label{eq:ssa_objfun}
\end{align}
which can be solved efficiently by using multiplicative updates with orthogonal matrices
parameterized as matrix exponentials of antisymmetric matrices \cite{PRL:SSA:2009,Plu05} \footnote{An efficient implementation of SSA may be downloaded free of charge at \url{http://www.stationary-subspace-analysis.org/toolbox}}.

\subsection{Finding the Most Non-Stationary Sources} 

In order to extract useful features for change-point detection, we would like to find 
the projection to the most non-stationary sources. However, the SSA algorithms
\cite{PRL:SSA:2009,HarKawWasBun10SSA} merely estimate the projection to the
most stationary sources, and choose the projection to the non-stationary sources to be
orthogonal to the found \s-projection, which means that all stationary contributions are 
projected out from the estimated \n-sources. The justification for this choice is that it maximizes 
the non-stationarity of the \n-sources in the case where the covariance between 
the true \n- and \s-sources is constant over time. This, however, may not always 
be the case: significant non-stationarity may well be contained in changing covariance
between \s- and \n-sources. In fact we observe this in our application to fault monitoring.
Thus, in order to find the most non-stationary sources, we also need to optimize 
the \n-projection. Before we turn to the optimization problem, let us first of all analyze 
the situation more formally.
 
\begin{figure}[ht]
 \begin{center}
  \includegraphics[width=87mm]{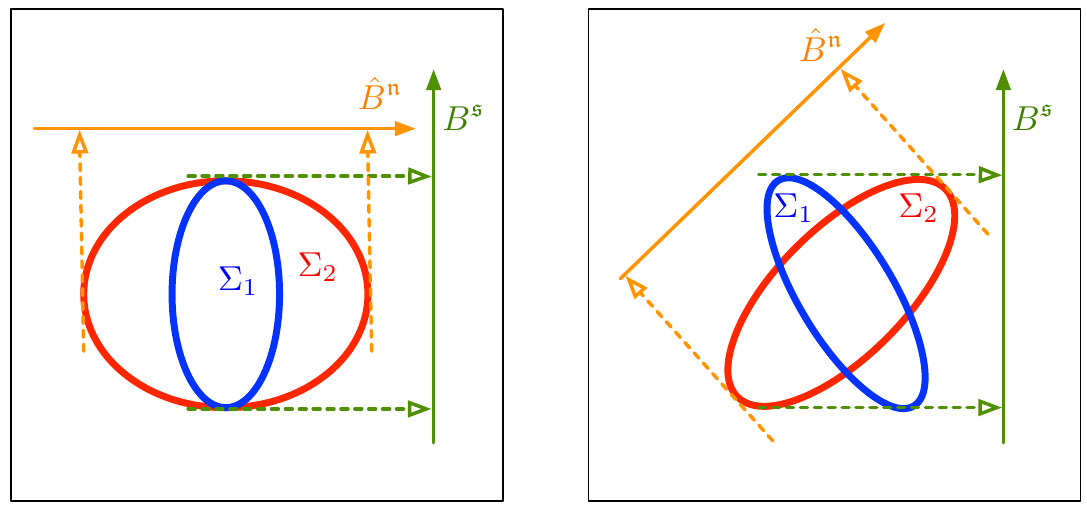}
  \caption{
  	The left panel shows two epoch covariance matrices $\Sigma_1$ and $\Sigma_2$ where the
	non-stationarity is confined to changes in the variance along one direction, hence the
	most non-stationary projection $\hat{B}^\n$ is orthogonal to the true stationary 
	projection $B^\s$. This is not the case in the situation depicted in the right panel: 
	here, the covariance of the two dimensions changes between $\Sigma_1$ and $\Sigma_2$, 
	so that we can find a non-stationary projection that is more non-stationary than 
	the orthogonal complement of the true \s-projection. 
    \label{fig:nstat_optim}
    }
 \end{center}
\end{figure}

We consider first a simple example where we have one stationary and one non-stationary source with 
corresponding normalized basis vectors $\| A^\s \| = 1$ and $\| A^\n \| = 1$ respectively, 
and let $\phi$ be the angle between the two spaces, i.e.~$\cos \phi = A^{\s \top} A^\n$. We will 
consider an arbitrary pair of epochs, $\mathcal{T}_1$ and $\mathcal{T}_2$, and show 
which projection $\hat{B}^\n$ maximizes the difference in mean $\Delta_\mu$ and variance $\Delta_\sigma$ 
between $\mathcal{T}_1$ and $\mathcal{T}_2$.

Let $X_1$ and $X_2$ be bivariate random variables modeling the distribution of 
the data in the two epochs respectively. According to the linear mixing model (Equation~\ref{eq:mixing_model}), 
we can write $X_1$ and $X_2$ in terms of the underlying sources,  
\begin{align*}
	X_1 & = A^\s X_s + A^\n X_{n_1} \\
	X_2 & = A^\s X_s + A^\n X_{n_2} 
\end{align*}
where the univariate random variable $X_s$ represents the stationary source and the two 
univariate random variables $X_{n_1}$ and $X_{n_2}$ model the non-stationary sources, 
in the epochs $\mathcal{T}_1$ and $\mathcal{T}_2$ respectively. Without loss of generality, we will 
assume that the true \s-projection $B^\s = (A^\n)^\perp$ is normalized, $\| B^\s \| = 1$. In order
to determine the relationship between the true \s-projection and the most non-stationary
projection, we write it in terms of $B^\s$ and $A^\n$,
\begin{align}
\label{eq:defnproj}
	\hat{B}^\n = \alpha B^\s + \beta A^{\n \top}, 
\end{align}
with coefficients $\alpha, \beta \in \R$ such that $\| \hat{B}^\n \| = 1$.
In the next step, we will observe which \n-projection maximizes the difference in 
mean $\Delta_\mu$ and covariance $\Delta_\sigma$ between the two epochs  $\mathcal{T}_1$ and $\mathcal{T}_2$. 
Let us first consider the difference in the mean of 
the estimated \n-sources, 
\begin{align*}
	\Delta_\mu = \E[ \hat{B}^\n X_1 ] - \E[ \hat{B}^\n X_2 ] =  \hat{B}^\n A^\n ( \E[X_{n_1}] -   \E[X_{n_2}] ) .
\end{align*}
This is maximal for $\hat{B}^\n A^\n = 1$, i.e.~when $\hat{B}^\n$ is orthogonal to $B^\s$.  
Thus, with respect to the difference in the mean, choosing the \n-projection $\hat{B}^\n$ to be orthogonal to the \s-projection is 
always optimal, irrespective of the type of distribution change between epochs.

Let us now consider the difference in variance $\Delta_\sigma$ of the estimated \n-sources
between epochs. This is given by, 
\begin{multline*}
	\Delta_\sigma = \Var[ \hat{B}^\n X_1 ] - \Var[ \hat{B}^\n X_2 ] =  \beta^2 ( \Var [ X_{\n_1} ] - \Var [ X_{\n_2} ] )
												\\ + 2  \left[ \alpha \cos \left(\phi + \frac{\pi}{2}\right) + \beta \cos \phi \right] \underbrace{( \Cov[ X_\s,  X_{\n_1} ] - \Cov[ X_\s,  X_{\n_2} ] )}_{= \Delta_{\sigma_{\s \n}}} .
\end{multline*}
Clearly, when there is no change in the covariance of the \s- and the \n-sources between the two epochs, 
i.e.~$\Delta_{\sigma_{\s \n}} = 0$, the difference $\Delta_\sigma$ is maximized for
 $\hat{B}^\n = (B^\s)^\perp$. See the left panel of Figure~\ref{fig:nstat_optim} for an example.
However, when the covariance between \s- and \n-sources does 
vary, i.e.~$| \Delta_{\sigma_{\s \n}}|>0$, the projection $(B^\s)^\perp$ is no longer 
the most non-stationary. To see this, consider the derivative of $\Delta_\sigma$ 
with respect to the $\alpha$ at $\alpha = 0$, 
\begin{align*}
	\partial \Delta_\sigma / \partial \alpha |_{\alpha=0} = 2  \cos \left(\phi + \frac{\pi}{2}\right) \Delta_{\sigma_{\s \n}} . 
\end{align*}
Since this derivate does not vanish, $\alpha = 0$ (see Equation~\ref{eq:defnproj}) is not an extremum when 
$|\Delta_{\s \n}|>0$, which means that the most non-stationary
projection is not orthogonal to the true \s-projection. This is the case in the right panel of
Figure~\ref{fig:nstat_optim}. 

Thus, in order to find the projection to the most non-stationary sources, we also need to
maximize the non-stationarity of the estimated \n-sources. To that end, we simply maximize the 
SSA objective function (Equation~\ref{eq:ssa_objfun}) for the \n-projection, 
\begin{align}
	\hat{B}^\n = \argmax_{B B^\top = I} \; \sum_{i=1}^N \left(
			- \log\det\esi^\n_i
			+ (\hat{\mu}^\n_i)^\top \emu^\n_i 	\right) ,
\label{eq:ssa_nobjfun}
\end{align}
where $\esi_i^\n = \hat{B}^\n \esi_i (\hat{B}^\n)^\top$ and $\hat{\mu}^\n_i =  \hat{B}^\n \emu_i$ for 
all epochs $1 \leq i \leq N$. 

\subsection{Relationship to Statistical Testing}

In this section we show that maximizing the SSA objective function to find the most 
non-stationary sources can be understood from a statistical testing point-of-view, 
in that it also maximizes the $p$-value for rejecting the null hypothesis that the estimated directions are stationary. 

More precisely, we maximize the $p$-value for a statistical hypothesis test that compares 
two models for the data: the null hypothesis $H_0$ that each epoch follows a standard normal 
distribution vs.~the alternative hypothesis $H_A$ that each epoch is Gaussian distributed with 
individual mean and covariance matrix. Let $X_1, \ldots, X_N$ be random variables modeling 
the distribution of the data in the $N$ epochs. Formally, the hypothesis can 
be written as follows.
\begin{align*}
\label{eq:hyp}
	& H_0 : X_1, \ldots, X_N \sim \mathcal{N}(0, I) \\
	& H_A : X_1 \sim \mathcal{N}(\mu_1, \Sigma_1), \ldots, X_N \sim \mathcal{N}(\mu_N, \Sigma_N)
\end{align*}
In other words, the statistical test tells us whether we 
should reject the simple model $H_0$ in favor of the more complex model $H_A$. This decision 
is based on the value of the test statistics, whose distribution is known under the null hypothesis 
$H_0$. Since $H_0$ is a special case of $H_A$ and since the parameter estimates are obtained by 
Maximum Likelihood, we can use the likelihood ratio test statistic $\Lambda$ \cite{Likelihood},
which is the ratio of the likelihood of the data under $H_0$ and $H_A$, where the parameters are
their maximum likelihood estimates. 

Let $\mathcal{X} \subset \R^{d_n}$ be the data set which is divided into 
$N$ epochs $\mathcal{T}_1, \ldots, \mathcal{T}_n$ and let $\emu^\n_1, \ldots, \emu^\n_N$ 
and $\esi^\n_1, \ldots, \esi^\n_N$ be the maximum likelihood estimates of the mean and covariance 
matrices of the estimated \n-sources respectively. Let $p_{\mathcal{N}}(x ; \mu, \Sigma)$ be the probability density 
function of the multivariate Gaussian distribution. The likelihood ratio test statistic is given by
\begin{align}
  \Lambda(\mathcal{X}) = - 2 \log \frac{ \prod_{x \in \mathcal{X}} p_{\mathcal{N}}(x ; 0, I)  } 
  { 
  	\prod_{i=1}^N \prod_{x \in \mathcal{T}_i} p_{\mathcal{N}}(x ; \emu^\n_i , \esi^\n_i) 
  }
\end{align}
which is approximately $\chi^2$ distributed with $\frac{1}{2} N d_n (d_n+3)$ degrees of freedom. 
Using the facts that we have set the average epoch's mean and covariance matrix to zero and 
the identity matrix respectively, i.e.~
\begin{align}
	\frac{1}{N} \sum_{i=1}^N \emu^\n_1  = 0 \hspace{0.5cm} \text{ and } \hspace{0.5cm} \frac{1}{N} \sum_{i=1}^N \esi^\n_i = I, 
\end{align}
the test statistic simplifies to, 
\begin{align}
\label{eq:teststat}
	\Lambda(\mathcal{X}) = -\frac{d_s}{2} N  + \frac{1}{2} \sum_{i=1}^N N_i \left( - \log \det \esi^\n_i  + \| \emu^\n_i \|^2 + \text{tr}(\esi^{\n}_i) \right),  
\end{align}
where $N_i = | \mathcal{T}_i |$ is the number of data points in the $i$-th epoch. If 
every epoch contains the same number of data points ($N_1 = \cdots = N_N$), then 
maximizing the SSA objective function (Equation~\ref{eq:ssa_nobjfun}) is equivalent 
to maximizing the test statistic (Equation~\ref{eq:teststat}) and hence minimizing the $p$-value 
for rejecting the simple (stationary) model for the data. 

As we will see in the application to fault monitoring (Section~\ref{sec:Real}), the $p$-value of this test furnishes a useful indicator of the number of informative
directions for change point detection. More specifically, we increase the number of candidate stationary sources until the test returns that they are significantly non-stationary.
As long as the $p-$value is large, we may safely conclude that these estimated stationary sources ($\hat{B}^\s{\mathbf s}(t)$) are stationary and that they may be removed without loss of informativeness for change point detection. In the specific case of change point detection, removing additional directions may lead to increases in performance as a result of the reduced dimensionality: this is of course a data dependent question.

\begin{figure}[ht]
 \begin{center}
  \includegraphics[width = 80mm]{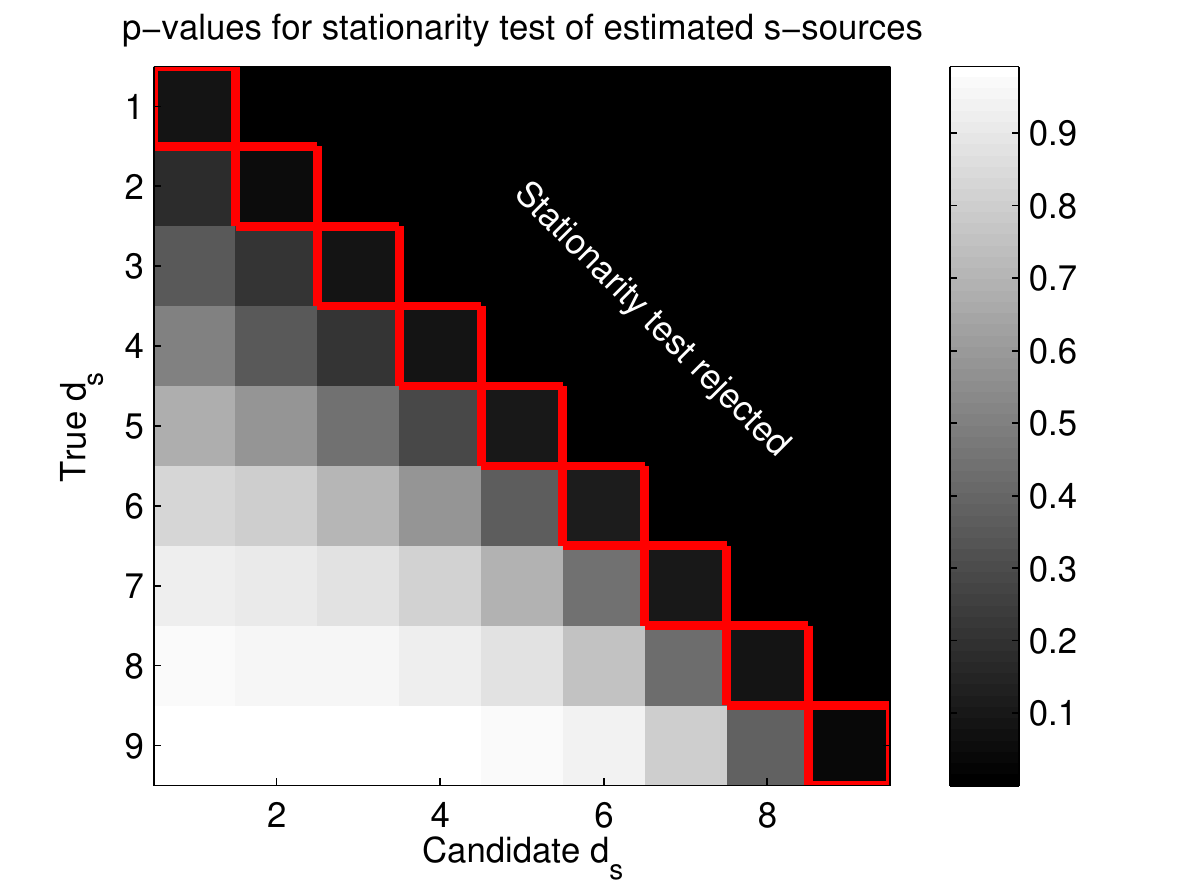}
  \caption{Average p-values obtained over 100 realizations of the dataset for each setting of the true $d_s$. The number of dimensions is $D=10$ and the $y$-axis displays the actual number for $d_s$.
  The $x-$axis displays the value for the parameter $d_s$ used to compute the stationary projection using SSA. The red box shows the decision made at the $p=0.01$ confidence level. The red box displays
  the choice made using this decision rule for choosing the parameter $d_s$ which occurs most often. The results show that the method picks the correct parameter on averages, over simulations.
     \label{fig:p_values}
    }
 \end{center}
\end{figure}

In Figure~\ref{fig:p_values} we illustrate the results of the procedure for choosing $d_s$, the number of stationary sources. We use the dataset described below in Section~\ref{sec:Sim}. The procedure for choosing $d_s$ is as follows: for each $\hat{d_s}$ from $1,\dots,D$, where $D$ is the dimensionality of the data set we compute the projection to the stationary sources. For each $\hat{d_s}$ we calculate the test statistic  $\Lambda(\mathcal{X})$. We then choose $d_s$ to be the largest such $d_s'$ such that we do not reject the null hypothesis, at the $p = \alpha$ confidence level, given in Equation~\ref{eq:hyp}.  We display the $p$-values obtained using SSA for a fixed value for simulated data's dimensionality $D=10$ and the number of stationary sources ranging from $d_1 = 1, \dots, 9$ and the chosen parameter ranging from $1, \dots, 9$.  The confidence level $\alpha = 0.01$ for rejection of the null hypothesis $H_0 = \text{\emph{The projected data is stationary}}$ returns the correct $d_s$ on average in all cases.

\section{Simulations}
\label{sec:Sim}

In this section we demonstrate the ability of SSA to enhance the segmentation performance of three change-point 
detection algorithms on a synthetic data setup. The algorithms are single linkage clustering with divergence (SLCD) \cite{Gower69SingleLinkage}  which uses the mean and 
covariance as test statistics, CUSUM \cite{Page:1954fk}, which uses a sequence of hypothesis tests and the Kohlmorgen/Lemm \cite{DistBased2}, which uses a kernel density measure and a hidden Markov model. For each segmentation algorithm we compare the performance of 
the baseline case in which the dataset is segmented without preprocessing, the case in which the data is 
preprocessed by projecting to a random subspace and the case in which the dataset is preprocessed 
using SSA. We compare performance with respect to the following schemes of parameter variation:

\begin{enumerate}
\item The dimensionality $D$ of the time series is fixed and $d_n$, the number of non-stationary sources is varied.
\item The number $d_n$ of non-stationary sources is fixed and $d_s$, the number of the stationary sources is varied.
\item $D$, $d_n$ and $d_s$ are fixed and the power $p$ between the changes in the non-stationary sources is varied.
\end{enumerate}

For two of the change point algorithms which we test, SLCD and Kohlmorgen/Lemm, all three parameter variation schemes are 
tested. For CUSUM the second scheme does not apply as the method is a univariate method.

For each setup and for each realization of the dataset we repeat segmentation on the raw dataset, 
the estimated non-stationary sources after SSA preprocessing for that dataset and on a $d_n$ 
dimensional random projection of the dataset. The random projection acts as a comparison measure for 
the accuracy of the SSA-estimated non-stationary sources for segmentation purposes. 

\begin{table}
\begin{center} 
\begin{tabular}{| c | c | c | c | c || c | c | c | } 
\hline 
Setup & $D$ & $d_n$ & $d_s$ & $p$ & SLCD & Kohl./Lemm & CUSUM \\
\hline \hline
(1) & \tickNo & \tickYes & \tickYes & \tickNo & Fi. \ref{fig:LinkageROC}, P. 1 & Fi. \ref{fig:KL_ROC}, P. 1 & Fi. \ref{fig:CUSUM_ROC}, P. 1 \\
(2) & \tickYes & \tickNo & \tickYes & \tickNo & Fi. \ref{fig:LinkageROC}, P. 2 & Fi. \ref{fig:KL_ROC}, P. 2 & Fi. \ref{fig:CUSUM_ROC}, P. 2 \\
(3) & \tickNo & \tickNo & \tickNo & \tickYes & Fi. \ref{fig:LinkageROC}, P. 3 & Fi. \ref{fig:KL_ROC}, P. 3 & Fi. \ref{fig:CUSUM_ROC}, P. 3 \\
\hline
\end{tabular} 
\caption{Overview of simulations performed and corresponding figures reporting the results. A tick denotes that the corresponding parameter was varied in the experiment. A cross denotes that the corresponding parameter is kept fixed. ("P." denotes panel within the respective figures and "Fi." denotes the figure.) }
\end{center}
\end{table}

\subsection{Synthetic Data Generation}

The synthetic data which we use to evaluate the performance of change point detection methods is generated as a linear mixture of stationary and non-stationary sources.
The data is further generated epoch wise: each epoch has fixed length and each dataset consists of a concatenation of epochs.
The $d$ stationary sources are distributed normally on each epoch according to $\Gauss(0,I_{d_s})$. The other $d_n$ (non-stationary) source signals $s^\n(t)$ are distributed according to the active model $k$ of this epoch; this active model is one of five Gaussian distributions $\G_k = \Gauss(0,\Sigma_k)$: the covariance $\Sigma_k$ is a diagonal matrix whose eigenvalues are chosen at random from five log-spaced values between $\sigma_1^2 = 1/p$ and $\sigma_5^2 = p$; thus five covariances, corresponding to the $\G_k$ of the Markov chain are then chosen in this way. The transition between models over consecutive epochs follows a Markov model with transition probabilities:
\begin{equation}
	P_{ij} = 
	\begin{cases} 
		0.9   & i=j \\ 
		0.025 & i\neq j .
	\end{cases}	
\end{equation}

\subsection{Performance Measure}
\label{sec:perform}
In our experiments we evaluate the algorithms based on an estimation of the area under the $ROC$ curves (AUC) across realizations of the dataset. The true positive rate (TPR) and false positive rate (FPR) are defined with respect to the fixed epochs with respect to which we generate the synthetic dataset; a changepoint may only occur between two such epochs of fixed length. Each of the changepoint algorithms, which we test, reports changes with respect to the same division into epochs as per the synthetic dataset: thus the TPR and FPR are well defined.

We use the AUC because it provides information relating to a range of TPR and FPR.  In signal detection the tradeoff achieved between TPR and FPR depends on operational constraints: cancer diagnosis procedures must achieve a high TPR perhaps at the cost of a higher than desirable FPR. Network intrusion detection, for instance, may need to compromise the TPR given the computational demands set by too high an FPR. In order to assess detection performance across all such requirements the AUC provides the most informative measure: all tradeoffs are integrated over.

More specifically, each algorithm is accompanied by a parameter $\tau$ which controls the trade off between TPR and FPR. For SLCD this is the number of clusters, for CUSUM this is the threshold set on the log likelihood ratio and for the Kohlmorgen/Lemm this is the parameter controlling how readily a new state is assigned to the model. 

\subsection{Single Linkage Clustering with Symmetrized Divergence Measure (SLCD)}

Single Linkage Clustering with a symmetrized distance measure is a simple algorithm for change point detection which has, however, the advantage of efficiency and of segmentation based on a parameter independent distance matrix (thus detection may be repeated for differing tradeoffs between TPR and FPR without reevaluating the distance measure).
In particular, segmentation based on Single Linkage Clustering \cite{Gower69SingleLinkage} computes a distance measure based on the covariance and mean over time windows to estimate the occurrence of changepoints:
the algorithm consists of the following three steps.
\begin{enumerate}
	\item The time series is divided into 200 epochs for 
				which we estimate the epoch-mean and epoch-covariance matrices 
				$\{ ( \hat{\boldsymbol \mu}_i, \hat{\Sigma}_i ) \}_{i=1}^{200}$.

	\item The dissimilarity matrix $D \in \R^{200 \times 200}$ between the epochs is computed 
				as the symmetrized Kullback-Leibler divergence $\KLD$ between the estimated distributions
				(up to the first two moments),			  
				\begin{align*}
					D_{ij} = 	\frac{1}{2} \KLD\left[ \Gauss(\hat{\boldsymbol \mu}_i, \hat{\Sigma}_i ) \; || \; \Gauss(\hat{\boldsymbol \mu}_j, \hat{\Sigma}_j )  \right] + \\ 
	\frac{1}{2} \KLD\left[ \Gauss(\hat{\boldsymbol \mu}_j, \hat{\Sigma}_j ) \; || \; \Gauss(\hat{\boldsymbol \mu}_i, \hat{\Sigma}_i )  \right],  
				\end{align*}
				where $\Gauss({\boldsymbol \mu}, \Sigma)$ is the Gaussian distribution.

	\item Based on the dissimilarity matrix $D$, Single Linkage Clustering \cite{Gower69SingleLinkage} 
				(with number of clusters set to $k=5$) returns an assignment of epochs to clusters such that 
				a changepoint occurs when two neighbouring epochs do not belong to the same cluster.

\end{enumerate}

\begin{figure*}[ht]
 \begin{center}
  \includegraphics[width = 120mm]{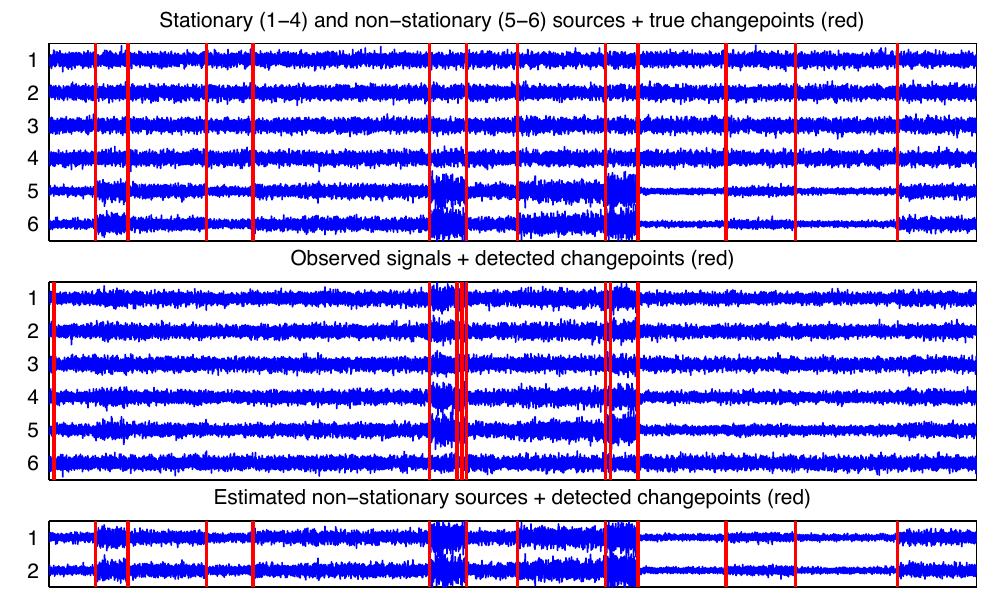}
  \caption{An Illustration of a case in which SSA significantly improves single linkage clustering with divergence: $d_s = 4$ (no. of stat. sources), $d_n = 2$ (no. non-stat.), $p = 2.3$ (power change in non-stat. sources). 
  The top panel displays the true decomposition into stationary and non-stationary sources with the true changepoints marked. The middle panel displays the
  changepoints which SLCD finds on the entire data set (sources mixed): clearly some changepoints are left undetected. The bottom panel displays the changepoints found by SLCD on the estimated non-stationary
  sources.
    \label{fig:seg_illustration}
    }
 \end{center}
\end{figure*}




\subsubsection{Results}

\begin{figure*}[ht]
 \begin{center}
  \includegraphics[width = 130mm]{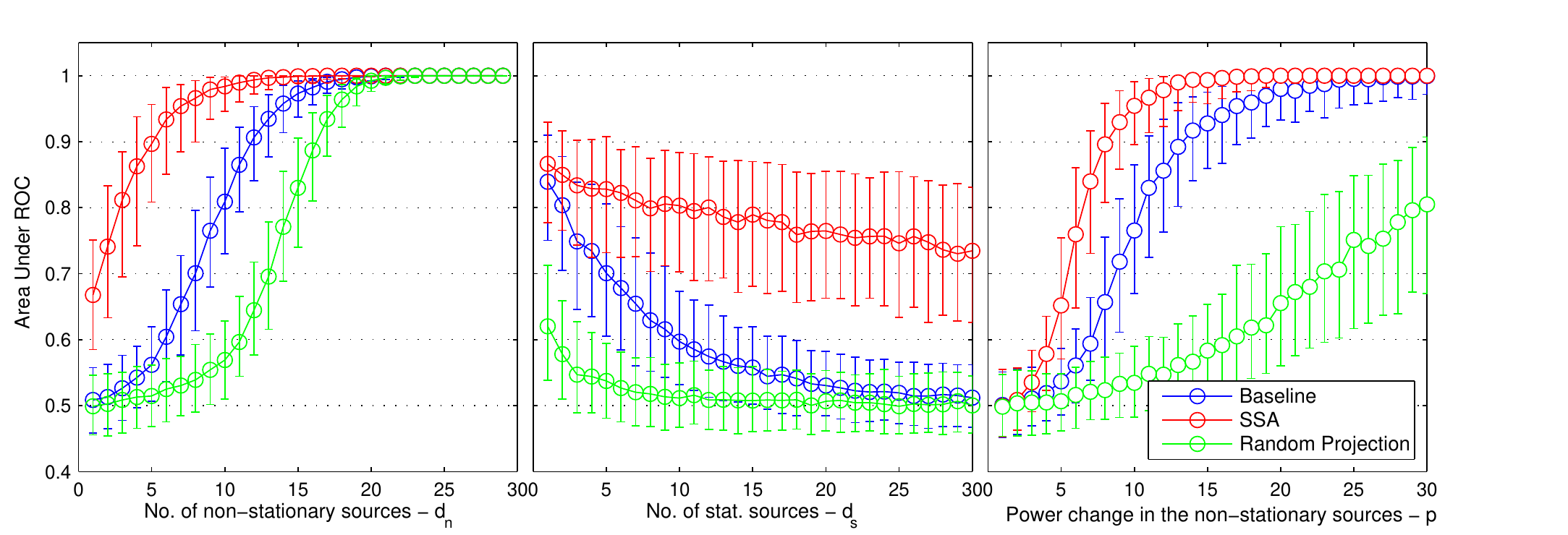}
  \caption{Results of the simulations for Single Linkage Clustering with Symmetrized Divergence (SLCD). The left panel displays the results for a fixed dimensionality of the time series, $D = 30$ and varying $d_n$, the number of stationary sources.The middle panel displays the results for a fixed number of non-stationary sources, $d_n = 2$ and varying  $d_s$, the number of stationary sources. The right panel displays the results for fixed $D = 20$, $d_s = 16$ and $d_n = 4$ and for varying $p$, the power change in the non-stationary sources. Each displays the results in terms of the area under the ROC curve computed as per Section \ref{sec:perform}.
  The error bars extend from the 25th to the 75th percentiles.
    \label{fig:LinkageROC}}
 \end{center}
\end{figure*}

The results of the simulations for varying numbers of non-stationary sources in  a dataset of 30 channels are shown in Figure~\ref{fig:LinkageROC} in the first panel. When the degree to which the changes are visible is lower, the SSA preprocessing significantly outperforms the baseline method, even for a small number of irrelevant stationary sources. 

The results of the simulations for a varying number of stationary dimensions with 2 non-stationary dimensions are displayed in Figure~\ref{fig:LinkageROC} in the second panel. For small $d$ the performance of the baseline and SSA preprocessing are similar: SSA's performance is more robust with respect to the addition of higher numbers of stationary sources, i.e. noise directions. The segmentations produced using SSA preprocessing continue to carry information relating to changepoints for $d_s = 30$, whereas, for $d \geq 12$, the baseline's AUC approaches $0.5$, which corresponds to the accuracy of randomly chosen segmentations.

The results of the simulations for varying power $p$ in the non-stationary sources with $D = 20$, $d_s = 16$ (no. of stat. sources) and $d_n = 4$ are displayed in Figure~\ref{fig:LinkageROC} in the third panel.
Both the performance of the Baseline and of the SSA preprocessing improves with increasing power change $p$. This effect is evident for lower $p$ for the SSA preprocessing
than for the baseline.

An illustration of a case in which SSA preprocessing significantly outperforms the baseline is displayed in Figure~\ref{fig:seg_illustration}. The estimated non-stationary sources exhibit a far clearer illustration of the changepoints than the full dataset: the corresponding segmentation performances reflect this fact.

\subsection{Weighted CUSUM for changes in variance}

In statistical quality control  CUSUM (or cumulative sum control chart) is a sequential analysis technique developed in 1954 \cite{Page:1954fk}. CUSUM is one of the most widely used and oldest methods for change point detection; the algorithm is an online method for change point detection based on a series of log-likelihood ratio tests.
Thus CUSUM algorithm detects a change in parameter $\theta$ of a process $p_\theta(y)$ \cite{Page:1954fk} and is asymtotically optimal when the pre-change and post-change parameters are known \cite{ChangePoint}.
For the case in which the target value of the changing parameter is unknown, the weighted CUSUM algorithm is a direct extension of CUSUM \cite{ChangePoint}, by integrating over a parameter interval, as follows. The following statistics ${\tilde \Lambda}_j^k$ constitutes likelihood ratios between the currently estimated parameter of the non-stationary process and differing target values (values to which the parameter may change), integrated over a measure $F$.
 
 \begin{equation}
{\tilde \Lambda}_j^k  = \Bigg{(} \int _{-\infty}^{\infty}\frac{p_{\theta_1}(y_j,...,y_k)}{p_{\theta_0}(y_j,...,y_k)}dF(\theta_1) \Bigg{)}
 \end{equation}  
 
 Here $y_j,...,y_k$ denote the timepoints lying inside a sliding window of length $k$ whereby $y_k$ indicates the latest time point received.
 The stopping time is then given as follows:
 
 \begin{equation}
 t_a = \mathrm{min}\{ k : \mathrm{max}\{ j \leq k : \mathrm{ln}({\tilde \Lambda}_j^k ) \geq h\} \}
 \end{equation}
 
The function $F$ serves as a weighting function for possible target values of the changed parameter. In principle the algorithm can thus be applied to multi-dimensional data. However, as per \cite{ChangePoint}, the extension of the CUSUM algorithm to higher dimensions is non-trivial, not just because integrating over possible values of the covariance is computationally expensive but also because various parameterizations can lead to the same likelihood function. Given this we test the effectiveness of the algorithm in computing one-dimensional segmentations. In particular we compare the segmentation performed on the one dimensional projection chosen by SSA with the best segmentation of all individual dimensions with respect to hit-rate on each trial. 
In accordance with \cite{ChangePoint} we choose $F$ to comprise a fixed uniform interval containing all possible values of the process's variance. We approximate the integral above as a sum over evenly spaced values on that interval. We approximate the stopping time by setting:
\begin{equation}
t_a \approx \mathrm{min}\{ k : \mathrm{ln}({\tilde \Lambda}_{k-W+1}^k ) \geq h\}
\end{equation}
The exact details of our implementation are as follows. Let $T$ be the number of data points in the data set $X$.

 \begin{enumerate}
	\item We set the window size $W$, the sensitivity constant $h$ and the current time step as $t_c=W+1$ and $\theta_0 = \text{var}(\{x_1,...,x_W\})$ and $\Theta = \{ \theta_1,\ldots,\theta_r \} = \{ c, c + b, c + 2b,\ldots, d\}$. 		
	\item ${\tilde \Lambda}_j^k = \frac{1}{b} \sum_{i = 1}^{r}\frac{p_{\theta_i}(y_j,\ldots,y_k)}{p_{\theta_0}(y_j,\ldots,y_k)}$
	\item If $\mathrm{ln}({\tilde \Lambda}_j^k) \geq h$ then a changepoint is reported at time $t_c$ and $t_c$ is updated so that $t_c = t_c + W$ and $\theta_0 = \text{var}(\{x_{t_c - W +1},\ldots,x_{t_c}\})$. We return to step 2.
	\item Otherwise if ${\tilde \Lambda}_j^k < h$ no changepoint is reported and $t_c = t_c + 1$. We return to step 2.
\end{enumerate}
 
 \subsubsection{Results}
  
\begin{figure*}[ht]
 \begin{center}
  \includegraphics[width = 120mm]{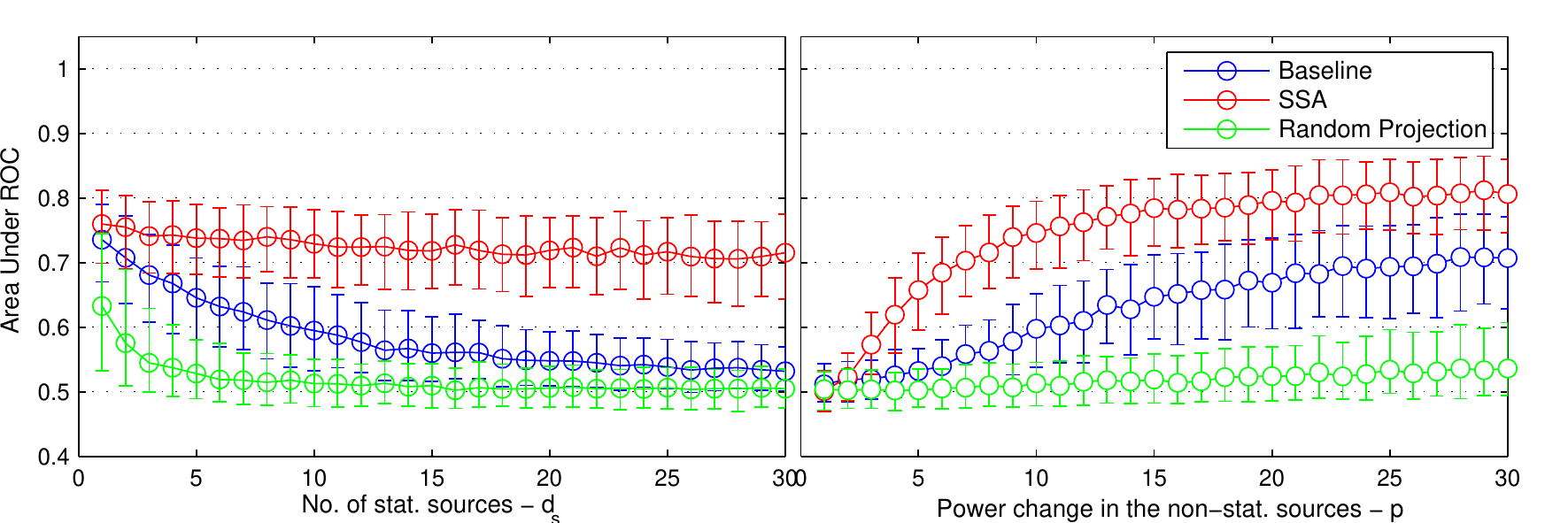}
  \caption{Results of the simulations for CUSUM. The left panel displays the results for a fixed number of non-stationary sources, $d_n = 1$ and varying  $d_s$, the number of stationary sources. The right panel displays the results for fixed $D = 16$ and $d = 15$ and for varying $p$, the power change in the non-stationary sources. Each displays the results in terms of the area under the ROC curve computed as per Section \ref{sec:perform}.
  The error bars extend from the 25th to the 75th percentiles.
    \label{fig:CUSUM_ROC}
    }
 \end{center}
\end{figure*}

 In Figure~\ref{fig:CUSUM_ROC}, in the left panel, the results for varying numbers of stationary sources are displayed. Weighted CUSUM with SSA preprocessing significantly outperforms the baseline for all values of D (dimensionality of the time series). Here we set $d_n=1$, the number of non-stationary sources, for all values of $d_s$, the number of stationary sources.
 
 In Figure~\ref{fig:CUSUM_ROC}, the right panel, the results for changes in the power change between ergodic sections $p$ are displayed for $D = 16$, $d_s=15$ and $d_n = 1$. SSA outperforms the baseline for all except very low values of $p$, the power level change, where all detection schemes fail. The simulations show that SSA represents a method for choosing a one dimensional subspace to render uni-dimensional segmentation methods applicable to higher dimensional datasets: the resulting segmentation method on the one dimensional derived non-stationary source will be simpler to parametrize and more efficient. If the true dimensionality of the non-stationary part is $d_n$ then no information loss should be observed.

%
 
\subsection{Kohlmorgen/Lemm Algorithm}

The Kohlmorgen/Lemm algorithm is a flexible non-parametric and multivariate method which may be applied in online and offline operation modes.
Distinctive about the Kohlmorgen/Lemm algorithm is that a kernel density estimator, rather than a simple summary statistic, is used to estimate the occurence of changepoints. In particular the algorithm is based on a standard Kernel Density Estimator with Gaussian kernels and estimation of the optimal segmentation based on a Hidden Markov Model  \cite{DistBased2}. More specifically if we estimate the densities on two arbitrary epochs $E_i,E_j$ of our dataset $X$ with Gaussian kernels then we can define a distance measure
$d$ between epochs via the $L2$-Norm yielding:

\begin{align*}
d(E_i,E_j) = \frac{1}{W^2 (4\pi \sigma^2)^{d/2}} \sum_{w,v = 0}^{W-1} \Bigg( \mathrm{exp}\left( - \frac{(Y_w - Y_v)^2}{4\sigma^2})\right)  \\  
 -  2\mathrm{exp}\left( - \frac{(Y_w - Z_v)^2}{4\sigma^2})\right) \\ + \mathrm{exp}\left( - \frac{(Z_w - Z_v)^2}{4\sigma^2})\right)\Bigg)
\end{align*}

The final segmentation is then based on the distance matrix generated between epochs calculated with respect to the above distance measure $d$.
As per the weighted CUSUM, it is possible to define algorithms whose sensitivity to distributional changes in reporting changepoints is related to the value of a parameter $C$: $C$ controls the probability of transitions to new states in the fitting of the hidden markov model. However, in \cite{Kohlmorgen:2003fk} it is shown that in the case when all changepoints are known then one can also derive an algorithm which returns exactly that number of changepoints: in simulations we evaluate the performance on the first variant over a full range of parameters to obtain an ROC curve. In addition we choose the parameter $\sigma$ according to the rule of thumb given in \cite{DistBased2}, which sets $\sigma$ proportional to the mean distance of each data point to its $D$ nearest neighbours, where $D$ is the dimensionality of the data: this is evaluated on a sample set. The exact implementation we test is based on the papers \cite{Kohlmorgen:2003fk} and \cite{DistBased2}.  The details are as follows:

\begin{enumerate}
\item The time series is divided into epochs 
\item A distance matrix is computed between epochs using kernel density estimation and the $L2$-norm as described above.
\item The estimated density on each epoch corresponds to a state of the Markov Model. So a state sequence is a sequence of estimated densities.
\item Finally, based on the estimated states and distance matrix, a hidden Markov model is fitted to the data and a change point reported whenever consecutive epochs have been fitted with differing states.
\end{enumerate}

\subsubsection{Results}

\begin{figure*}[ht]
 \begin{center}
  \includegraphics[width = 130mm]{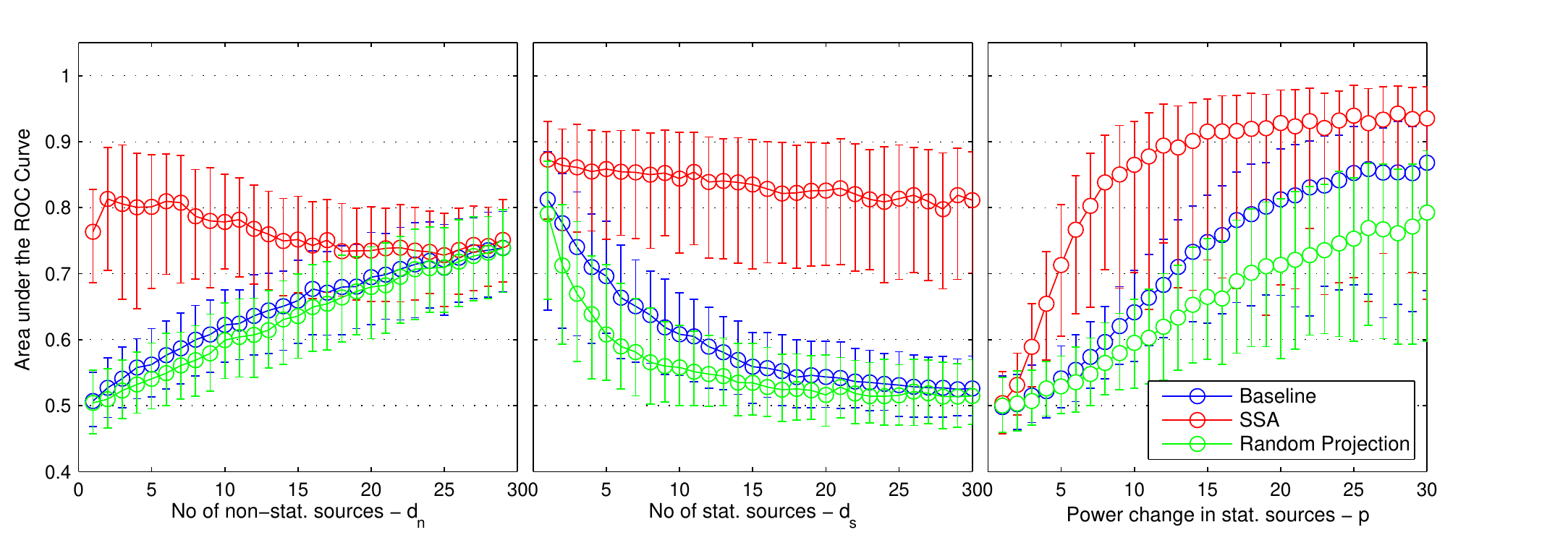}
  \caption{Results of the simulations for the Kohlmorgen/Lemm.The left panel displays the results for a fixed dimensionality of the time series, $D = 30$ and varying $d_n$, the number of stationary sources.The middle panel displays the results for a fixed number of non-stationary sources, $d_n = 2$ and varying  $d_s$, the number of stationary sources. The right panel displays the results for fixed $D = 20$, $d_s = 16$, $d_n=4$ and for varying $p$, the power change in the non-stationary sources. Each displays the results in terms of the area under the ROC curve computed as per Section \ref{sec:perform}. The error bars extend from the 25th to the 75th percentiles.
    \label{fig:KL_ROC}
    }
 \end{center}
\end{figure*}

SSA preprocessing improves the segmentation obtained using the Kohlmorgen/Lemm algorithm for all 3 setups of the dataset.
In particular:
the area under the ROC (AUC) for varying $d_s$ and fixed $D$ are displayed in Figure~\ref{fig:KL_ROC}, in the first panel, with $D=30$.
The area under the ROC (AUC) for varying $d_s$ and fixed $d_n$ are displayed in Figure~\ref{fig:KL_ROC}, in the second panel, with $d_n=2$.
The area under the ROC (AUC) for varying power change in the non-stationary sources $p$ and fixed $D,d$ are displayed in Figure~\ref{fig:KL_ROC}, in the third panel, with p ranging between 1.1 and 4.0 at increments of 0.1.
Of additional interest is that for varying $d_n$ and fixed $D$ the performance of segmentation with SSA  preprocessing is superior for higher values of $d_s$: this implies that the improvement of change point detection of the Kohlmorgen/Lemm algorithm due to the reduction in dimensionality to the informative estimated n-sources outweighs the difficulty of the problem of estimating the n-sources in the presence of a large number of noise dimensions.

%
%

\section{Application to Fault Monitoring}
\label{sec:Real}

In this section we apply our feature extraction technique to fault monitoring. The dataset consists of multichannel 
measurements of machine vibration. The machine under investigation is a pump, driven by an electromotor. The incoming shaft is reduced in speed by 
two delaying gear-combinations (a gear-combination is a combination of 
driving and a driven gear). Measurements are repeated for two identical machines, where the first shows a progressed pitting in both gears, and the second machine is virtually fault free. The rotating speed of the driving shaft is measured with a tachometer. \footnote{The dataset can be downloaded free of charge at \url{http://www.ph.tn.tudelft.nl/~ypma/mechanical.html}.}

\begin{figure*}[!h]
\centering
\includegraphics[width = 122mm]{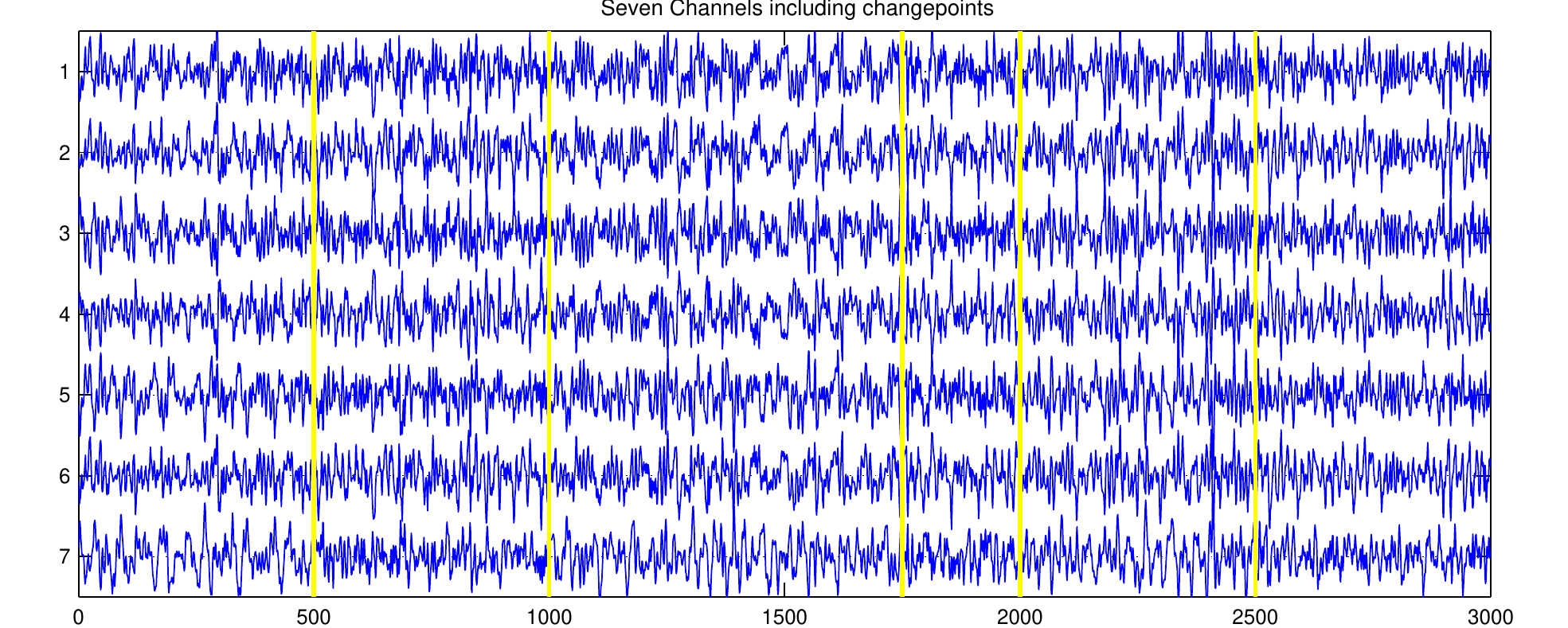}
\caption{Pump Dataset: the machine under investigation is a pump, driven by an electromotor. The measurements made are of machine vibration at seven sensors. The data alternates between two conditions: normal functionality and pitting in both gears. \label{fig:CLOSERLOOKDATA}}
 \end{figure*}

The pump data set is semi-synthetic insofar as we juxtapose non-temporally consecutive sections of data between the two pump conditions. Sections of data from the first and second machine are spliced randomly (with respect to the time axis) together to yield a dataset with 10,000 time points in seven channels.
An illustration of the dataset is displayed in Figure~\ref{fig:CLOSERLOOKDATA}.
                                                                                
\subsection{Setup}

We preprocessed with SSA using a division of the dataset into 30 equally sized epochs and $d$ estimated non-stationary sources, for $d_s$, the no. of stationary sources ranging between $1$ and $6$, where $D=7$ is the dimensionality of the dataset: 
subsequently we ran the KL algorithm on both the preprocessed and raw data using a window size of $W=50$ and a separation of 50 datapoints between non-overlapping epochs.


%
 
\subsection{Parameter Choice}
To select the parameter, $d_s$ (and thus $d_n = D-d_s$), the number of stationary sources, we use the following scheme: the measure of stationarity over which we optimize for SSA and SSA is given by the loss function in equation (2).
For each $d = \text{dim}(V^s)$ we compute the estimated projection to the stationary sources using SSA on the first half of the data available and computed this loss function on the estimated stationary sources on the second half and compared the result to the values of the loss function obtained on the dataset obtained by randomly permuting the time axis. This random permutation should produce, on average, a set of approximately stationary sources regardless of non-stationarity present in the estimated stationary sources for that $d$. In addition a measure of the information relating to non-stationarity lost in choosing the number of stationary sources to be $d_s$, the Baseline-Normalized Integral Stationary Error (BNISE), can be defined as followed:
\begin{equation}
\text{BNISE}(d) = \sum_{d'<d} \frac{L_{d'}(\hat{A}^{-1},X) - \mathbf{E}_{X'}(L_{d'}(\hat{A}^{-1}),X')}{\sigma_{X'}(L_{d'}(\hat{A}^{-1},X'))}
\end{equation}
Where $L_{d'}(\hat{A}^{-1},X)$ denotes the loss function given in equation~\ref{eq:ssa_objfun} on the original dataset with stationary parameter $r$ and  $L_{r}(\hat{A}^{-1},X')$ the same measure on a random permutation $X'$ of the same dataset.

\subsection{Results}

\begin{figure*}[!h]
 \centering
 \includegraphics[width = 110mm]{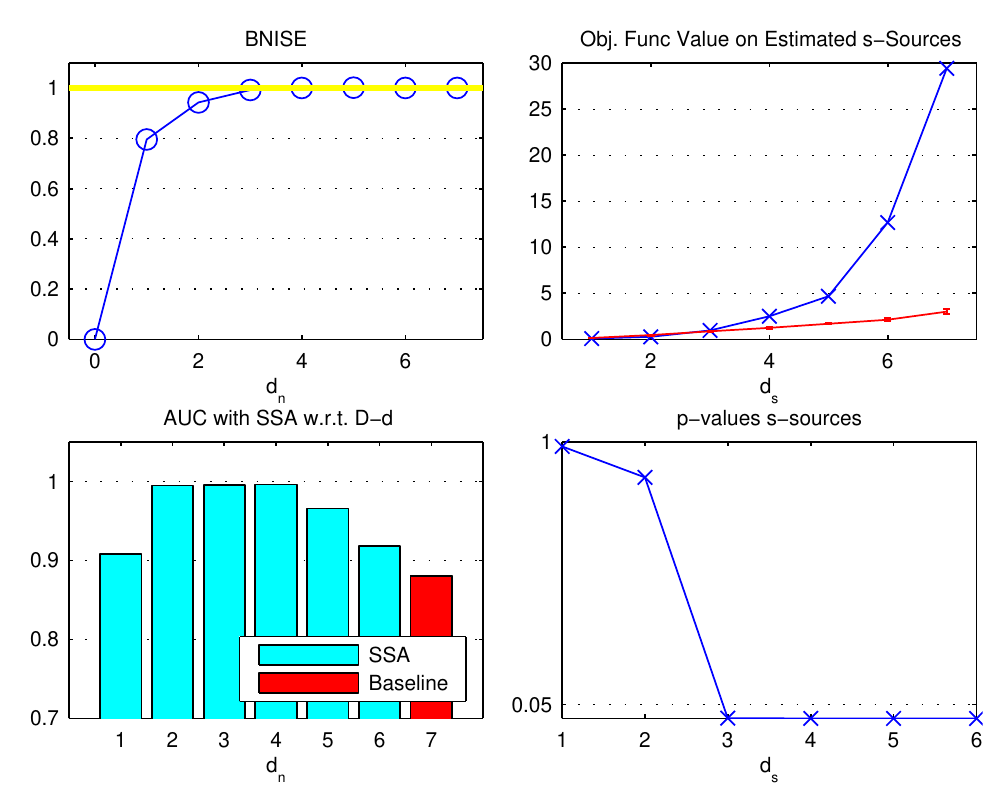}
 \caption{Pump Dataset: schemes for selecting the parameter $d_s$. Top left: the measure BNISE for increasing values of $d_n$. Top right the value of
 the error function as compared to randomly generated data. Bottom left: the AUC performances for various values of $d_n$. Bottom left: $p$-values
 on the estimated $s$-sources.\label{fig:ParameterPlots}}
 \end{figure*}

\begin{figure}[!h]
 \centering
 \includegraphics[width = 40mm]{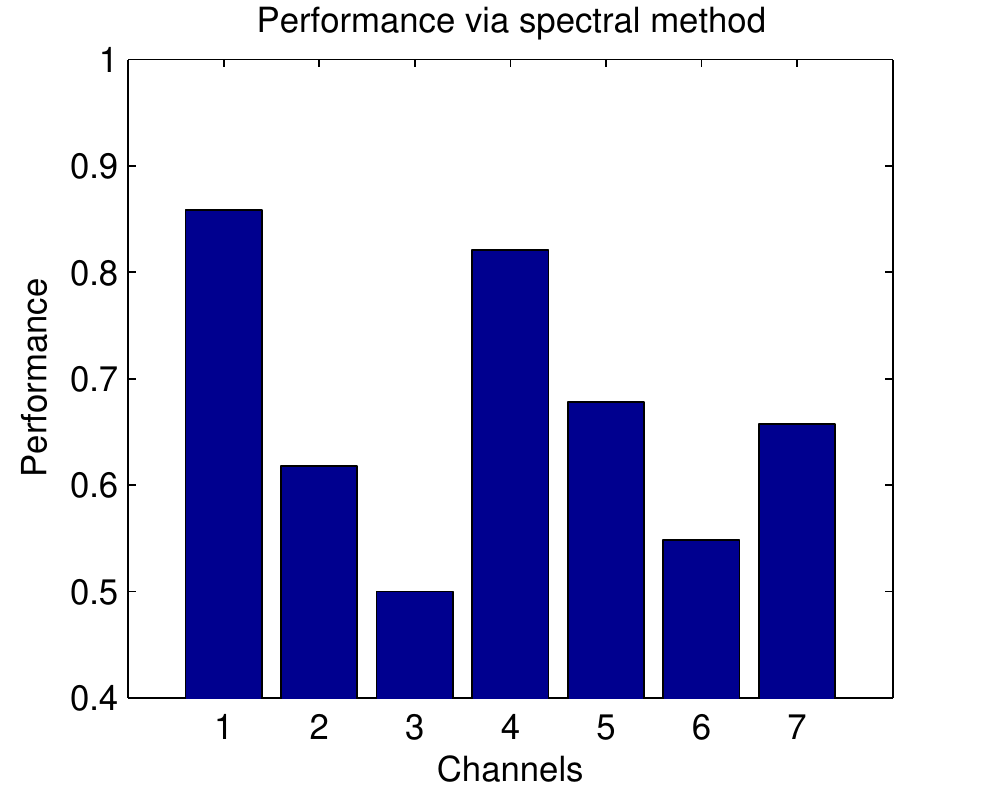}
 \caption{Pump Dataset: performance on spectograms computed on individual channels of the datatset. Each spectrogram is computed with a window length of 50 datapoints and overlap of 49 datapoints.
 7 frequency band windows are used to compute a timeseries of size $7 \times 10,000$.
  \label{fig:freq_pump}
}
 \end{figure}

The results of this scheme and the segmentation are given in Figure~\ref{fig:ParameterPlots}. For $d_s = 6$ we observe a clearly visible difference between the expected loss function value due to small sample sizes and the loss function value present in the estimated stationary sources. Similarly, looking at the p-values, we observe that for $d_s=1,2$ we do not reject the hypothesis that the estimated $\s-sources$ are stationary, whereas for higher values of $d_s$ we reject this hypothesis. This implies that $d_n \geq 5$. To test the effectiveness of this scheme, segmentation is evaluated for SSA preprocessing at all possible values of $d_s$.
The AUC values obtained using the parameter choices $d_s = 1,\ldots,6$ for SSA preprocessing as compared to the baseline case are displayed in Figure~\ref{fig:ParameterPlots}. An increase in performance with SSA preprocessing is robust, as measured by the AUC values, with respect to varying choices for the parameter $d_s$ as long as $d_s$ is not chosen $\leq 2$.
Note that, although, for the dataset at hand, there exists information relating to changepoints in the frequency spectrum taken over time, this information cannot be used to bring the baseline method onto
a par with preprocessing with SSA. We display the results in figure~\ref{fig:freq_pump} for comparison. Here, segmentation based on a 7-dimensional spectrogram based on each individual channel of the dataset is computed. The best performance over channels, for segmentation on each of these spectrograms is lower than the worst performance achieved on the entire dataset without using spectral information, with or without SSA.

\begin{figure*}[!h]
 \centering
 \includegraphics[width = 130mm]{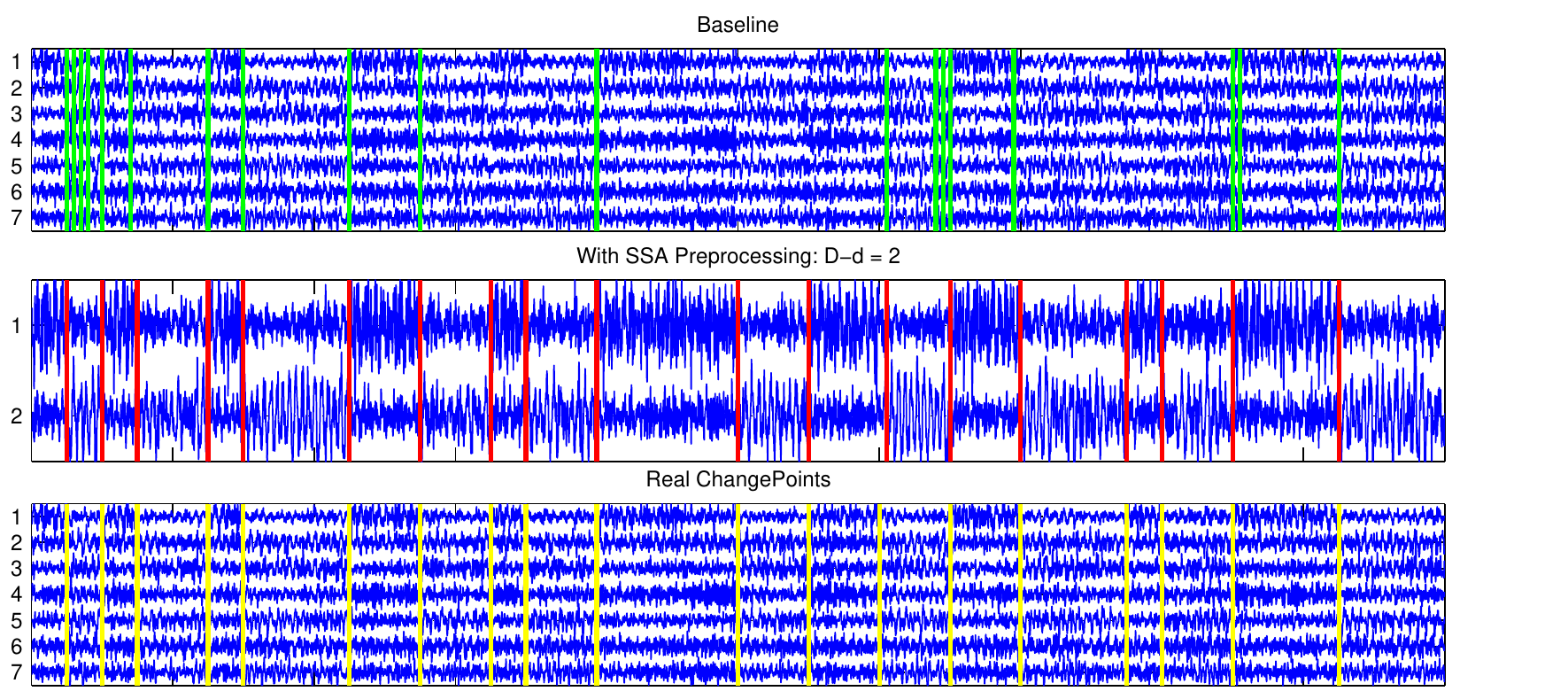}
 \caption{Pump Dataset: all segmentations are computed using Kohlmorgen/ Lemm with the number of changepoints $N$ specified \label{fig:PumpPicture}. The baseline corresponds to segmentation
 without SSA preprocessing. The middle panel displays segmentation with SSA preprocessing. The bottom panel displays the real changepoints superimposed over the raw dataset.
}
 \end{figure*}

\section{Conclusion}
\label{sec:conclusion}

Unsupervised segmentation and identification of time series is a hard problem even in the univariate case and has received considerable attention in science and industry due to its broad applicability that ranges from process control and finance to biomedical data analysis.

In high dimensional segmentation problems, different subsystems of the multivariate
time series may exhibit clearer and more informative signals for segmentation than others. The present 
contribution has harvested this property by decomposing the overall system into stationary and 
non-stationary parts by means of SSA and using the non-stationary subsystem to determine the segmentation. 

Intuitively segmentation can be understood as clustering in a function space (in which the estimated potentials reside) and, as shown in this paper, SSA contributes by choosing the most appropriate function space, which is most informative for the purpose of segmentation. 

This generic approach is shown to yield excellent results in simulations, illustrating the novel framework for segmentation made available by SSA. We expect that the proposed dimensionality reduction will 
be useful on a wide range of datasets, because the task of discarding irrelevant stationary information is 
independent of the dataset-specific distribution within the informative non-stationary subspace. Moreover, 
the SSA preprocessing is a highly versatile tool because it can be combined with any subsequent segmentation 
method. 

Applications made along the same lines as in the present contribution are effective only when the non-stationary part of the data is visible in the mean and covariances.
The present method may be thus made applicable to general datasets whose changes consist in the spectrum or temporal domain of the data by computing the score function as a further preprocessing step \cite{ChangePoint}. Future work will also focus on computing the projection to the non-stationary sources directly for data whose non-stationarity is more prominent in the spectrum than in the mean and covariance over time.

\bibliographystyle{plain}
{\small
\bibliography{stable,segmentation,ssa_pubs}}




\end{document}

%% file: ssa_defs.tex
\usepackage{bm}
\usepackage{amsmath}
\usepackage{amssymb}  

\usepackage{dsfont}

\usepackage[normalem]{ulem}

\usepackage{dsfont}

\newcommand{\R}{\ensuremath{\mathds{R}}}		

\newcommand{\Gauss}{\ensuremath{\mathcal{N}}}	

\newcommand{\s}{\ensuremath{\mathfrak{s}}}		
\newcommand{\n}{\ensuremath{\mathfrak{n}}}		

\DeclareMathOperator*{\E}{\mathds{E}}					
\DeclareMathOperator*{\Var}{Var}				
\DeclareMathOperator*{\Cov}{Cov}				
\DeclareMathOperator*{\argmin}{argmin}		    
\DeclareMathOperator*{\argmax}{argmax}		    

\newcommand{\KLD}{D_{\text{KL}}}				
\newcommand{\esi}{\hat{\Sigma}}					
\newcommand{\emu}{{\ensuremath{\hat{\mu}}}}	

\newcommand{\NN}{\mathbb{N}}
